\pdfoutput=1

\documentclass[11pt]{article}

\usepackage[]{EMNLP2022}

\usepackage{times}
\usepackage{multirow}
\usepackage{amsmath}
\usepackage{amssymb}
\usepackage{graphicx}
\usepackage{caption}
\usepackage{subcaption}
\usepackage{xcolor}
\usepackage{float}
\usepackage{booktabs}
\usepackage{natbib}
\usepackage{latexsym}
\usepackage[textsize=scriptsize]{todonotes}
\usepackage{algorithm}
\usepackage{algpseudocode}
\usepackage{algorithmicx}
\usepackage{hyperref}
\usepackage{xr}
\usepackage{bbm}
\usepackage{enumitem}
\usepackage{listings}

\renewcommand{\algorithmiccomment}[1]{\bgroup\hfill\small//~#1\egroup}

\newcommand{\dataname}[0]{\textsc{ClaimDecomp}}

\usepackage[T1]{fontenc}

\usepackage[utf8]{inputenc}

\usepackage{microtype}

\usepackage{inconsolata}

\title{Generating Literal and Implied Subquestions\\to Fact-check Complex Claims}

\author{Jifan Chen \ \ \ \ \ \ \ Aniruddh Sriram \ \ \ \ \ \ \ Eunsol Choi \ \ \ \ \ \ \ Greg Durrett \\
  Department of Computer Science \\
  The University of Texas at Austin \\
  \texttt{jfchen@cs.utexas.edu}}
 
\begin{document}
\maketitle
\begin{abstract}



Verifying political claims is a challenging task, as politicians can use various tactics to subtly misrepresent the facts for their agenda. Existing automatic fact-checking systems fall short here, and their predictions like ``half-true'' are not very useful in isolation, since it is unclear which parts of a claim are true or false. In this work, we focus on decomposing a complex claim into a comprehensive set of yes-no subquestions whose answers influence the veracity of the claim. We present \dataname{}, a dataset of decompositions for over 1000 claims. Given a claim and its verification paragraph written by fact-checkers, our trained annotators write subquestions covering both explicit propositions of the original claim and its implicit facets, such as additional political context that changes our view of the claim's veracity. We study whether state-of-the-art pre-trained models can learn to generate such subquestions. Our experiments show that these models generate reasonable questions, but predicting implied subquestions based only on the claim (without consulting other evidence) remains challenging. Nevertheless, we show that predicted subquestions can help identify relevant evidence to fact-check the full claim and derive the veracity through their answers, suggesting that claim decomposition can be a useful piece of a fact-checking pipeline.\footnote{We release our code and dataset: \url{https://jifan-chen.github.io/ClaimDecomp}}

\end{abstract}

\section{Introduction}

Despite a flurry of recent research on automated fact-checking \cite{wang-2017-liar,Rashkin2017TruthOV,volkova2017separating,ferreira2016emergent,popat2017truth,fake2018tschiatschek}, we remain far from building reliable fact-checking systems \cite{Nakov2021AutomatedFF}. This challenge motivated us to build more explainable models so the explanations can at least help a user interpret the results \cite{atanasova-etal-2020-generating-fact}. However, such purely extractive explanations do not necessarily help users interpret a model's {reasoning process}. An ideal explanation should do what a human-written fact-check does: systematically dissect different parts of the claim and evaluate their veracity. 

\begin{figure}[t]
\centering
\includegraphics[width=0.5\textwidth]{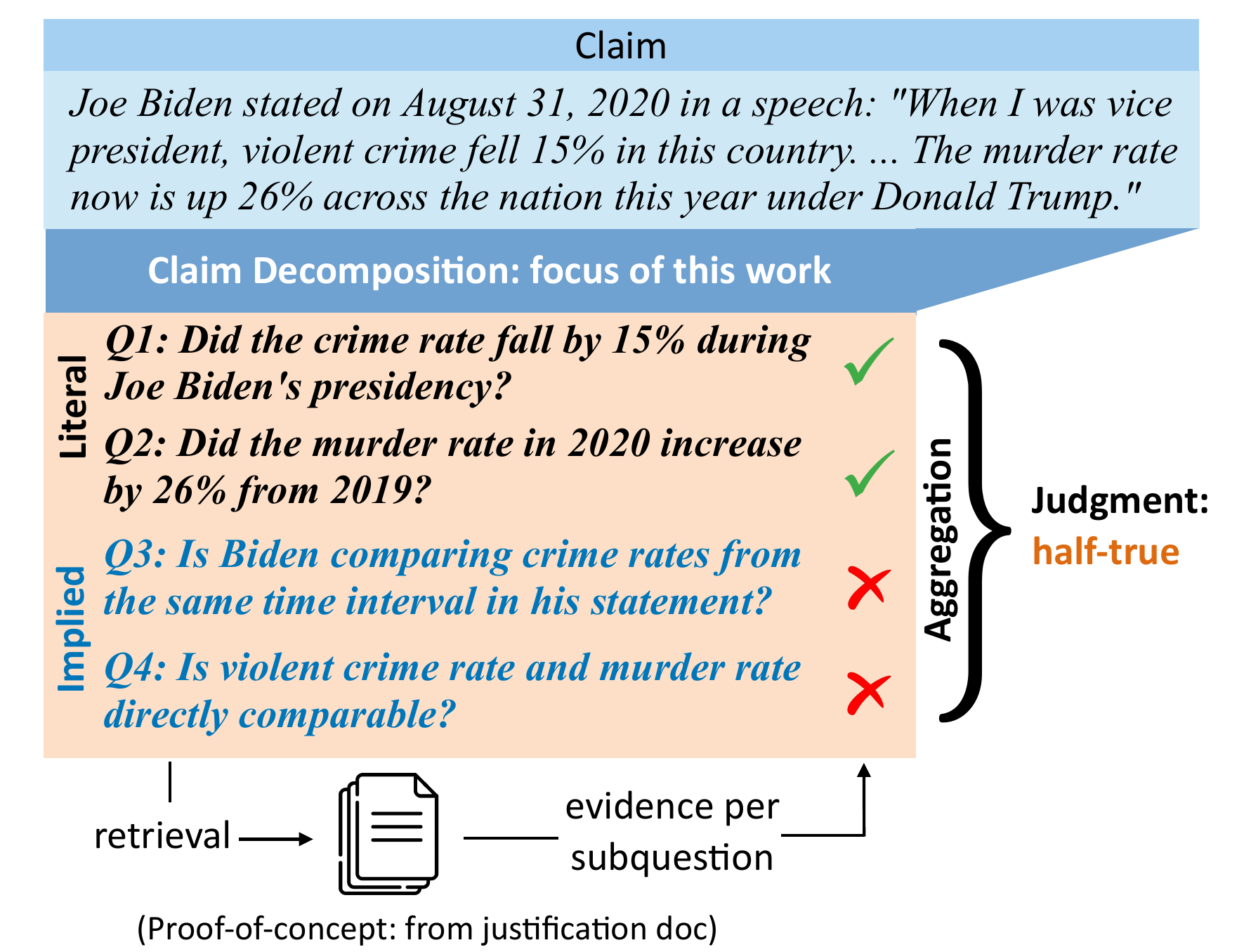}
\caption{An example claim decomposition: the top two subquestions follow explicitly from the claim and the bottom two represent implicit reasoning needed to verify the claim. We can use the decomposed questions to retrieve relevant evidence (Section~\ref{sec:evidence-retrieval}), and aggregate the decisions of the sub-questions to derive the final veracity of the claim (Section~\ref{sec:aggregate-sub-questions}).}
    \label{fig:intro}
\end{figure}

\begin{figure*}[t]
\centering
\includegraphics[width=\textwidth]{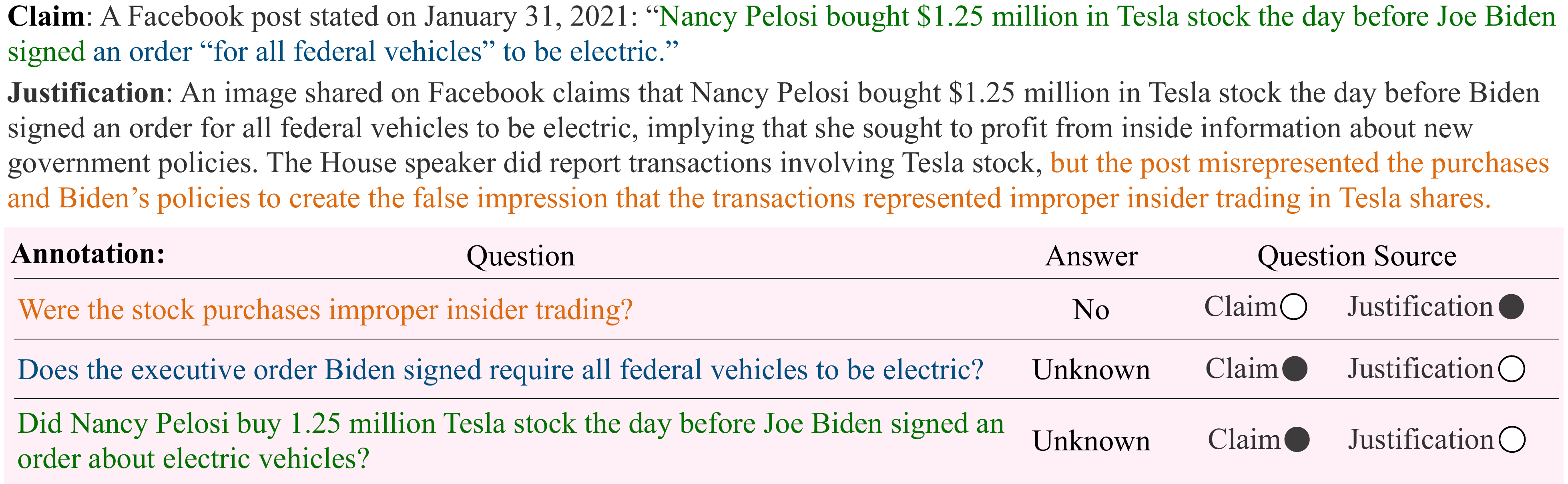}
\caption{An example of our annotation process. The annotators are instructed to write a set of subquestions, give binary answers to them, and attribute them to a source. If the answer cannot be decided from the justification paragraph, ``Unknown'' is also an option. The question is either based on the claim or justification, and the annotators also select the relevant parts (color-coded in the figure) on which the question is based.}
    \label{fig:pseudo-annot-interface}
\end{figure*}

We take a step towards explainable fact-checking with a new approach and accompanying dataset, \dataname{}, of decomposed claims from PolitiFact. Annotators are presented with a claim \emph{and} the justification paragraph written by expert fact-checkers, from which they annotate a set of yes-no subquestions that give rise to the justification. These subquestions involve checking both the explicit and implicit aspects of the claim (Figure~\ref{fig:intro}).

Such a decomposition can play an important role in an interpretable fact verification system. First, the subquestions provide a comprehensive explanation of how the decision is made: in Figure~\ref{fig:intro}, although the individual statistics mentioned by Biden are correct, they are from different time intervals and not directly comparable, which yields the final judgment of the claim as ``half-true''. We can estimate the veracity of a claim using the decisions of the subquestions (Section~\ref{sec:aggregate-sub-questions}). Second, we show that decomposed subquestions allow us to retrieve more relevant paragraphs from the verification document than using the claim alone (Section~\ref{sec:evidence-retrieval}), since some of the subquestions tackle implicit aspects of a claim. We do not build a full pipeline for fact verification in this paper, as there are other significant challenges this poses, including information which is not available online or which needs to be parsed out of statistical tables \cite{singh-et-al-2021}. Instead, we focus on showing how these decomposed questions can fit into a fact-checking pipeline through a series of proof-of-concept experiments. 


Equipped with \dataname\ dataset, we train a model to generate decompositions of complex political claims. We experiment with pre-trained sequence-to-sequence models~\cite{raffel2020exploring}, generating either a sequence of questions or a single question using nucleus sampling \cite{holtzman-2020-nucleus} over multiple rounds. This model can recover 58\% of the subquestions, including some implicit subquestions. To summarize, we show that decomposing complex claims into subquestions can be learned with our dataset, and reasoning with such subquestions can lead improve evidence retrieval and judging the veracity of the whole claim. 


\begin{table*}[t]
\small
\centering
\begin{tabular}{ l c c | c || c c c | c c }
\toprule
       &     \# unique  & \# tokens &  avg. \# subquestions  & \multicolumn{3}{c}{Answer \%} & \multicolumn{2}{|c}{Source \%}  \\
 Split &  claims & per claim  & in single annotation       & Yes & No & Unknown               & Justification & Claim    \\
\midrule
Train & 800&  33.4 &2.7 & 48.9 & 45.3 & 5.8 & 83.6 & 16.4 \\
Validation & 200  & 33.8 & 2.7 & 48.3 & 44.8 & 6.9 & 79.0 & 21.0  \\
Validation-sub & 50  & 33.7 & 2.9 & 45.2 & 47.8 & 7.0 & 90.4 & 9.6 \\
Test & 200  &   33.2 &2.7 & 45.8 & 43.1 & 11.1 & 92.1 & 7.9  \\
\bottomrule
\end{tabular}
\caption{Statistics of the \dataname{} dataset. Each claim is annotated by two annotators, yielding a total of 6,555 subquestions. The second column blocks (Answer \% and Source \%) report the statistics at the subquestion level; Source \% denotes the percentage of subquestions based on the text from the justification or the claim.}
\label{tab:dataset-statistics}
\end{table*}

\section{Motivation and Task}

Facing the complexities of real-world political claims, simply giving a final veracity to a claim often fails to be persuasive~\cite{guo2022survey}. To make the judgment of an automatic fact-checking system understandable, most previous work has focused on generating \emph{justifications} for models' decisions. \citet{popat-etal-2018-declare, shu2019defend, lu-li-2020-gcan} used attention weights of the models to highlight the most relevant parts of the evidence, but these only deal with explicit propositions of a claim. \citet{ahmadi2019explainable, gad2019exfakt} used logic-based systems to generate justifications, yet the systems are often based on existing knowledge graphs and are hard to adapt to complex real-world claims. \citet{atanasova-etal-2020-generating-fact} treated the justification generation as a summarization problem in which they generate a justification paragraph according to some relevant evidence. Even so, it is hard to know which parts of the claim are true and which are not, and how the generated paragraph relates to the veracity. 

What is missing in the literature is a better intermediate representation of the claim: with more complex claims, explaining the veracity of a whole claim at once becomes more challenging. Therefore, we focus on decomposing the claim into a \textbf{minimal} yet \textbf{comprehensive} set of yes-no subquestions, whose answers can be aggregated into an inherently explainable decision. As the decisions to the subquestions are explicit, it is easier for one to spot the discrepancies between the veracity and the intermediate decisions.

\paragraph{Claims and Justifications} Our decomposition process is inspired by fact checking documents written by professional fact checkers. In the data we use from PolitiFact, each \textbf{claim} is paired with a \textbf{justification paragraph} (see Figure~\ref{fig:pseudo-annot-interface}) which contains the most important factors on which the veracity made by the fact-checkers is based. Understanding \emph{what questions are answered in this paragraph} will be the core task our annotators will undertake to create our dataset. However, we frame the claim decomposition task (in the next section) without regard to this justification document, as it is not available at test time.


\paragraph{Claim Decomposition Task}
\label{sec:task}

We define the task of complex claim decomposition. Given a claim $c$ and the context $o$ of the claim (speaker, date, venue of the claim), the goal is to generate a set of $N$ yes-no subquestions $\mathbf{q} = \{q_1,q_2,...q_N\}$. The \textbf{set} of subquestions should have the following properties:

\begin{itemize}
    \item \textbf{Comprehensiveness:} The questions should cover as many aspects of the claim as possible: the questions should be sufficient for someone to judge the veracity of the claim.

    \item \textbf{Conciseness:} The question set should be as minimal as is practical and not contain repeated questions asking about minor, correlated variants seeking the same information.
 
\end{itemize}

\noindent An individual subquestion should also exhibit: 

\begin{itemize}
    \item \textbf{Relevance:} The answer to subquestion should help a reader determine the veracity of the claim. Knowing an answer to a subquestion should change the reader's belief about the veracity of the original claim (Section~\ref{sec:aggregate-sub-questions}). 
    \item \textbf{Fluency / Clarity:} Each subquestion should be clear, fluent, and grammatically correct (Section~\ref{sec:inter-annotator-agreement}). 
\end{itemize}
We do not require subquestions to stand alone~\cite{Choi2021DecontextualizationMS}; they are instead interpreted with respect to the claim and its context.

\paragraph{Evaluation Metric}
We set the model to generate the target number of subquestions, which matches the number of subquestions in the reference, guaranteeing a concise subquestion set. Thus, we focus on measuring the other properties with reference-based evaluation. Specifically, given an annotated set of subquestions and an automatically predicted set of subquestions, we assess \textbf{recall}: how many subquestions in the reference set are covered by the generated question set? 
A subquestion in the reference set is considered as being recalled if it is \textbf{semantically equivalent} to one of the generated subquestions by models.\footnote{There are cases where one generated question covers several reference questions, e.g., treating the whole claim as a question, in which case we only consider one of the reference questions to be recalled.} Our notion of equivalence is nuanced and contextual: for example, the following two subquestions are considered semantically equivalent: \emph{``Is voting in person more secure than voting by mail?''} and \emph{``Is there a greater risk of voting fraud with mail-in ballots?''}. We manually judge the question equivalence, as our experiments with automatic evaluation metrics did not yield reliable results (details in Appendix~\ref{sec:claim_decomp_eval}).

\section{Dataset Collection}

\paragraph{Claim / Verification Document Collection}
We collect political claims and corresponding verification articles from PolitiFact.\footnote{\url{https://www.politifact.com/}} Each article contains one justification paragraph (see Figure~\ref{fig:pseudo-annot-interface}) which states the most important factors on which the veracity made by the fact-checkers is based. Understanding what questions are answered in this paragraph will be the core annotation task. Each claim is classified as one of six labels: \emph{pants on fire} (most false), \emph{false}, \emph{barely true}, \emph{half-true}, \emph{mostly true}, and \emph{true}. We collect the claims from top 50 PolitiFact pages for each label, resulting in a total of 6,859 claims. 

A claim like \emph{``Approximately 60,000 Canadians currently live undocumented in the USA.''} hinges on checking a single statistic and is less likely to contain information beyond the surface form. Therefore, we mainly focus on studying complex claims in this paper. To focus on complex claims, we filter claims with 3 or fewer verbs. We also filter out claims that do not have an associated justification paragraph. After the filtering, we get a subset consisting 1,494 complex claims. 




\paragraph{Decomposition Annotation Process}
Given a claim paired with the justification written by the professional fact-checker on PolitiFact, we ask our annotators to reverse engineer the fact-checking process: generate yes-no questions which are answered in the justification. As shown in Figure~\ref{fig:pseudo-annot-interface}, for each question, the annotators also (1) give the answer; (2) select the relevant text in the justification or claim that is used for the generation (if any). The annotators are instructed to cover as many of the assertions made in the claim as possible without being overly specific in their questions. 

This process gives rise to both \textbf{literal questions}, which follow directly from the claim, and \textbf{implied questions}, which are not necessarily as easy to predict from the claim itself. These are not attributes labeled by the annotators, but instead labels the authors assign post-hoc (described in Section~\ref{sec:decomposition-comparison}).

\begin{table}[t]
\small
\centering
\begin{tabular}{ c  c c c  }
\toprule
  & \textsc{all qs} & \textsc{more qs} & \textsc{fewer qs}  \\
\midrule
 \% of unmatched Qs & 18.4 & 26.1 & 8.5 \\
\bottomrule
\end{tabular}
\caption{Inter-annotator agreement assessed by the percentage of questions for which the semantics cannot be matched to the other annotator's set. We name the question set containing more questions as \textsc{more qs} and the other one as \textsc{less qs}. \textsc{all qs} is the average of \textsc{more qs} and \textsc{less qs}. }
\label{tab:question-decomposition-agreement}
\end{table}

We recruit 8 workers with experience in literature or politics from the freelancing platform Upwork to conduct the annotation. Appendix~\ref{sec:question-annotation-workflow} includes details about the hiring process, workflow, as well as instructions and the UI.


\paragraph{Dataset statistics and inter-annotator agreement} \label{sec:inter-annotator-agreement}
Table~\ref{tab:dataset-statistics} shows the statistics of our dataset. We collect two sets of annotations per claim to improve subquestion coverage. We collect a total of 6,555 subquestions for 1,200 claims. Most of the questions arise from the justification and most of the questions can be answered by the justification. In addition, we randomly sample 50 claims from the validation set for our human evaluation in the rest of this paper. We name this set \textbf{Validation-sub}. 

Comparing sets of subquestions from different annotators is nontrivial: two annotators may choose different phrasings of individual questions and even different decompositions of the same claim that end up targeting the same pieces of information. Thus, we (the authors) manually compare two sets of annotations to judge inter-annotator agreement: given two sets of subquestions on the same claim, the task is to identify questions for which the semantics are not expressed by the other question \emph{set}. If no questions are selected, it means that the two annotators show strong agreement on what should be captured in subquestions. Example annotations are shown in Appendix~\ref{sec:inter-annotator-agreement-example}.

We randomly sample 50 claims from our dataset and three of the authors conduct the annotation. The authors agree on this comparison task reasonably, with a Fleiss' Kappa~\cite{fleiss1971measuring} value of 0.52. The comparison results are shown in Table~\ref{tab:question-decomposition-agreement}. On average, the semantics of 18.4\% questions are not expressed by the other set. This demonstrates the \textbf{comprehensiveness} of our set of questions: only a small fraction is not captured by the other set, indicating that independent annotators are not easily coming up with distinct sets of questions. Because most questions are covered in the other set, we view the agreement as high. A simple heuristic to improve comprehensiveness further is to prefer the annotator who annotated more questions. If we consider the fraction of unmatched questions in the \textsc{fewer qs}, we see this drops to 8.5\%.\footnote{Merging two annotations results in many duplicate questions and deduplicating these without another round of adjudication is cognitively intensive. We opted not to do this due to the effectiveness of simply taking the larger set of questions.} Through this manual examination, we also found that annotated questions are overall concise, fluent, clear, and grammatical.

\section{Automatic Claim Decomposition}
\label{sec:auto_decomp}

The goal is to generate a subquestion set $\mathbf{q}$ from the input claim $c$, the context $o$, and the target number of subquestions $k$. 

\begin{figure}[t]
\centering
\includegraphics[width=0.42\textwidth]{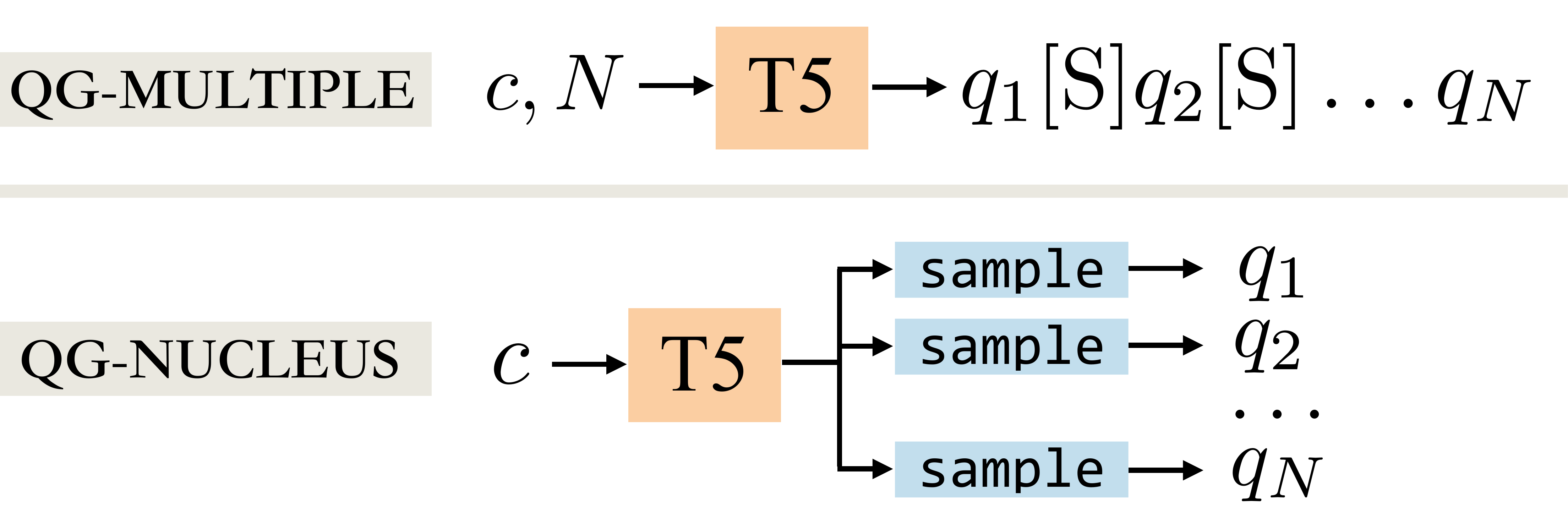}
\caption{Illustration of our two question generators. \textsc{qg-multiple} generates all questions as a sequence while \textsc{qg-nucleus} generates one question at a time through multiple samples.}
    \label{fig:question-generator}
\end{figure}

\paragraph{Models}
We fine-tune a T5-3B~\cite{raffel2020exploring} model to automate the question generation process under two settings: \textsc{qg-multiple} and \textsc{qg-nucleus} as shown in Figure~\ref{fig:question-generator}. Both generation methods generate the same number of subquestions, equal to the number of subquestions generated by an annotator.

\begin{table}[t]
\small
\centering
\begin{tabular}{ l  c c c c c c c}
\toprule
Model & R-all & R-literal & R-implied \\
\midrule
\textsc{qg-multiple} &  0.58 & 0.74 & 0.18	 \\
\textsc{qg-nucleus} & 0.43 & 0.59 & 0.11  \\

\textsc{qg-multiple-justify} & 0.81 & 0.95 & 0.50 \\
\textsc{qg-nucleus-justify} & 0.52 & 0.72 & 0.18 \\
\bottomrule
\end{tabular}
\caption{Human evaluation results on the Validation-sub set (N=146). R-all denotes the recall for all questions; R-literal and R-implied denotes the recall for the literal questions and the implied questions respectively. }

\label{tab:qg-eval}
\end{table}

\paragraph{\textsc{qg-multiple}} We learn a model $P(\mathbf{q} \mid c, o)$ to place a distribution over sets of subquestions given the claim and output. The annotated questions are concatenated by their annotation order to construct the output.

\paragraph{\textsc{qg-nucleus}} We learn a model $P(q \mid c, o)$ to place a distribution over single subquestions given the claim and output. For training, each annotated subquestion is paired with the claim to form a \emph{distinct} input-output pair. At inference, we use nucleus sampling to generate questions.  See Appendix~\ref{sec:model-details} for training details.

We also train these generators in an oracle setting where the justification paragraph is appended to the claim to understand how well the question generator does with more information. We denote the two oracle models as \textsc{qg-multiple-verify} and \textsc{qg-nucleus-verify} respectively. 


\begin{table}[t]
\small
\centering
\begin{tabular}{ c | c | c c c }
\toprule
Question Type & \# Questions & R1-P & R2-P & RL-P \\
\midrule
 Literal & 2.15 & 0.56 & 0.30 & 0.47 \\
 Implied  & 1.02 & 0.28 & 0.09 & 0.22 \\
\bottomrule
\end{tabular}
\caption{Number of questions of each type per claim and their lexical overlap with the claim measured by ROUGE-1, ROUGE-2, and ROUGE-L precision (how many $n$-grams in the question are also in the claim).}
\label{tab:annotation-lexical-overlap}
\end{table}

\begin{figure*}[t]
\centering
\includegraphics[width=\textwidth]{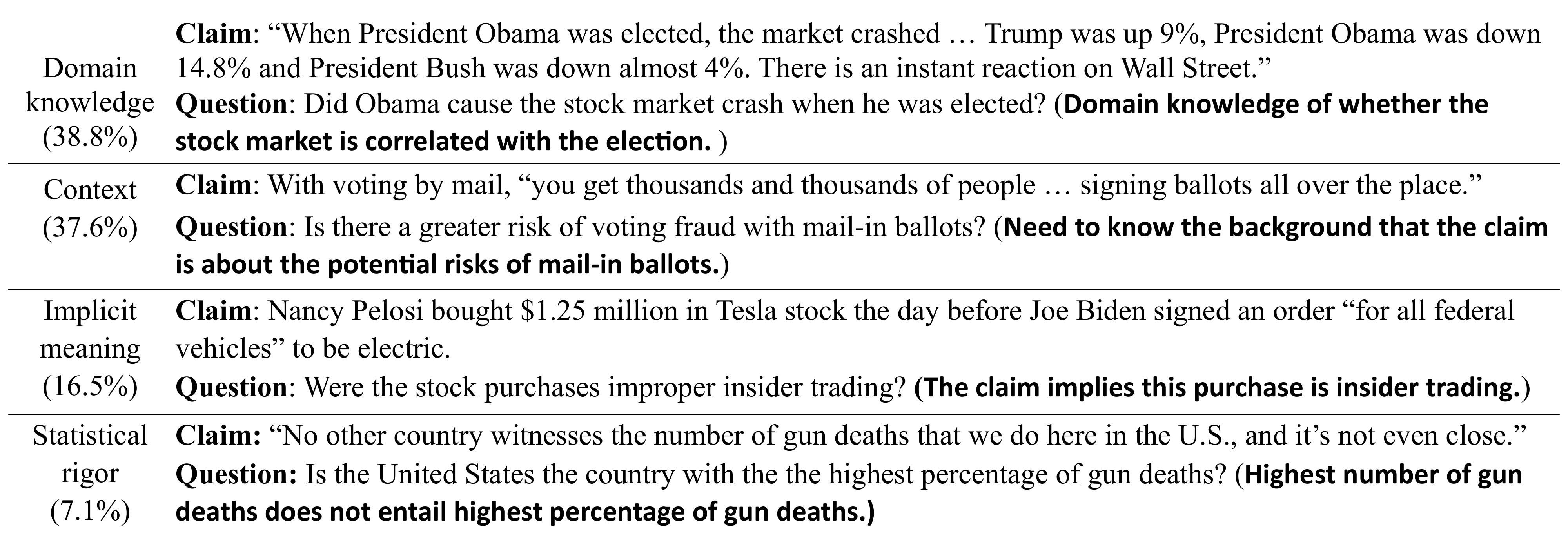}
\caption{Four types of reasoning needed to address subquestions with their proportion (left column) and examples (right column). It shows that a high proportion of the questions need either domain knowledge or related context.}
\label{fig:annot-analysis}
\end{figure*}

\paragraph{Results}\label{exp:t5-decomposition}
All models are trained on the training portion of our dataset and evaluated on the Validation-sub set. One of the authors evaluated the recall of each annotated subquestion in the generated subquestion set. 
The results are shown in Table~\ref{tab:qg-eval}. We observe that \textbf{most of the literal questions can be generated while only a few of the implied questions can be recovered.} Generating multiple questions as a single sequence (\textsc{qg-multiple}) is more effective than sampling multiple questions (\textsc{qg-nucleus}). Many questions generated from \textsc{qg-nucleus} are often slightly different but share the same semantics. We see that more than 70\% of the literal questions and 18\% of the implied questions can be generated by the best \textsc{qg-multiple} model. By examining the generated implied questions, we find that most of them belong to the \textbf{domain knowledge} category in Section~\ref{sec:decomposition-comparison}. 

Some questions could be better generated if related evidence were retrieved first, especially for questions of the \textbf{context} category (Section~\ref{sec:decomposition-comparison}). The \textsc{qg-multiple-justify} model can recover most of the literal questions and half of the implied questions. Although this is an oracle setting, it shows that when given proper information about the claim, the T5 model can achieve much better performance. We discuss this retrieval step more in Section~\ref{sec:limitations}.

\paragraph{Qualitative Analysis} While our annotated subquestion sets cover most relevant aspects of the claim, we find some generated questions are good subquestions that are missing in our annotated set, though less important. For example, for our introduction example shown in Figure~\ref{fig:intro}, the \textsc{qg-nucleus} model generates the question ``\emph{Is Trump responsible for the increased murder rate?}'' Using the question generation model in collaboration with humans might be a promising direction for more comprehensive claim decomposition. See Appendix~\ref{sec:qualitative-analysis-qg} for more examples.

\section{Analyzing Decomposition Annotations} \label{sec:decomposition-comparison}
In this section, we study the characteristics of the annotated questions. We aim to answer: (1) How many of the questions address implicit facets of the claim, and what are the characteristics of these? (2) How do our questions differ from previous work on question generation for fact checking~\cite{fan-etal-2020-generating}? (3) Can we aggregate subquestion judgments for the final claim judgment?

\subsection{Subquestion Type Analysis}
We (the authors) manually categorize 285 subquestions from 100 claims in the development set into two disjoint sets: \emph{literal} and \emph{implied}, where \emph{literal} questions are derived from the surface information of the claim -- whether a question can be posed by only given the claim, and \emph{implied} questions are those that need extra knowledge in order to pose.

Table~\ref{tab:annotation-lexical-overlap} shows basic statistics about these sets, including the average number of subquestions for each claim and lexical overlap between subquestions and the base claims, evaluated with ROUGE precision, as one subquestion can be a subsequence of the original claim. On average, each claim contains one implied question which represents the deeper meaning of the claim. These implied questions overlap less with the claim.

We further manually categorize the implied questions into the following four categories, reflecting what kind of knowledge is needed to pose them (examples in Figure~\ref{fig:annot-analysis}). Two authors conduct the analysis over 50 examples and the annotations agree with a Cohen's Kappa~\cite{cohen1960coefficient} score of 0.74.

\noindent \textbf{Domain knowledge}\quad The subquestion seeks domain-specific knowledge, for example asking about further steps of a legal or political process.

\noindent \textbf{Context}\quad The subquestion involves knowing that broader context is relevant, such as whether something is broadly common or the background of the claim (political affiliation of the politician, history of the events stated in the claim, etc). 

\noindent \textbf{Implicit meaning}\quad The subquestion involves unpacking the implicit meaning of the claim,  specifically anchored to what the speaker's intent was.

\noindent \textbf{Statistical rigor}\quad The subquestion involves checking over-claimed or over-generalized statistics (e.g., the highest raw count is not the highest per capita).

 Most of the implied subquestions require either domain knowledge or context about the claim, reflecting the challenges behind automatically generating such questions.

\begin{figure*}[t]
\centering
\includegraphics[width=\textwidth]{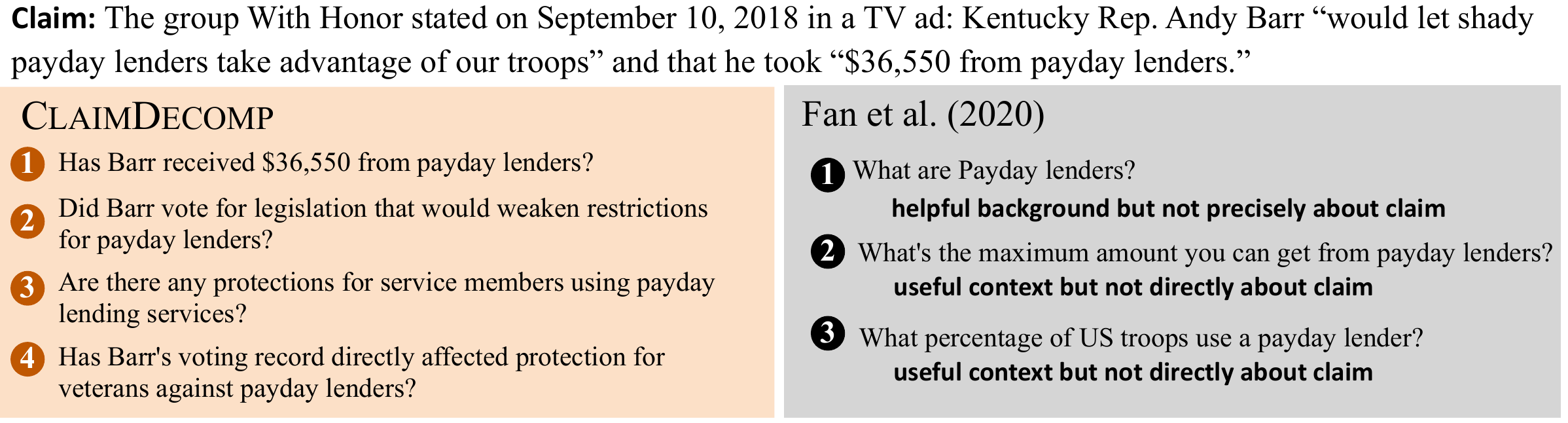}
\caption{Comparison between our decomposed questions with QABriefs~\cite{fan-etal-2020-generating}. In general, our decomposed questions are more comprehensive and relevant to the original claim.}
    \label{fig:annot-analysis-vs-fan-et-al}
\end{figure*}

\subsection{Comparison to QABriefs} \label{sec:question-comparison}

Our work is closely related to the QABriefs dataset~\cite{fan-etal-2020-generating}, where they also ask annotators to write questions to reconstruct the process taken by professional fact-checkers provided the claim and its verification document. 

While sharing similar motivation, we use a significantly different annotation process than theirs, resulting in qualitatively different sets of questions as shown in Figure~\ref{fig:annot-analysis-vs-fan-et-al}. We notice: (1) Their questions are less comprehensive, often missing important aspects of the claim. (2) Their questions are broader and less focused on the claim. We instructed annotators to provide the \textbf{source} of the annotated subquestions from either claim or verification document. For example, questions like ``\emph{What are Payday lenders?}'' in the figure will not appear in our dataset as the justification paragraph does not address such question.
\newcite{fan-etal-2020-generating} dissuaded annotators from providing binary questions; instead, they gather answers to their subquestions after the questions are collected. We focus on binary questions whose verification could help verification of the full claim. See Appendix~\ref{sec:compare-to-fan-et-al-more-examples} for more examples of the comparison.





\begin{table}[t]
\small
\centering
\begin{tabular}{ l |  c c c }
\toprule
 & mean & std & \# examples  \\
\midrule
 QABriefs~\cite{fan-etal-2020-generating} & 2.88 & 1.20 & 210 \\
 Ours & 3.60 & 1.19 &  210 \\
 
 \midrule
 $p$-value & \multicolumn{3}{c}{$\leq$ 0.0001} \\
 mean diff & \multicolumn{3}{c}{0.72} \\
 95\% CI & \multicolumn{3}{c}{0.48 - 0.97} \\ 
\bottomrule
\end{tabular}
\caption{Results from user study on helpfulness (rated 1-5) of a set of generated subquestions for claim verification. We conduct a t-test over the collected scores. }
\label{tab:annotation-analysis-vs-fan}
\end{table}

\begin{table}[t]
\small
\centering
\begin{tabular}{ l |  c c c }
\toprule
   & Macro-F1 & Micro-F1 & MAE \\
\midrule
 Question aggregation & 0.30 & 0.29 & 1.05 \\
 Question aggregation* & 0.46 & 0.45 & 0.73 \\
 
 Random (label dist) & 0.16 & 0.18 & 1.68 \\
 Most frequent & 0.06 & 0.23 & 1.31 \\
\bottomrule
\end{tabular}
\caption{Claim classification performance of our question aggregation baseline vs.~several baselines on the development set. MAE denotes mean absolute error.}
\label{tab:annotation-analysis-label-aggregation}
\end{table}

\paragraph{User Study} 
To better quantify the difference, we also conduct a user study in which we ask an annotator to rate how useful a set of questions (without answers) are to determine the veracity of a claim. On 42 claims annotated by both approaches, annotators score sets of subquestions on a Likert scale from 1 to 5, where 1 denotes that knowing the answers to the questions does not help at all and 5 denotes that they can accurately judge the claim once they know the answer. We recruit annotators from MTurk. We collect 5-way annotation for each example and conduct the t-test over the results. The details can be found in Appendix~\ref{sec:user-study-interface}. 

Table~\ref{tab:annotation-analysis-vs-fan} reports the user study results. Our questions achieve a significantly higher relevance score compared to questions from QABriefs. This indicates that we can potentially derive the veracity of the claim from our decomposed questions since they are binary and highly relevant to the claim.

\subsection{Deriving the Veracity of Claims from Decomposed Questions} \label{sec:aggregate-sub-questions}
Is the veracity of a claim sum of its parts? We estimate whether answers to subquestions can be used to determine the veracity of the claim.

We predict a veracity score $\hat{v} = \frac{1}{N} \sum_{i=1}^N \mathbbm{1}[a_i=1]$ equal to the fraction of subquestions with yes answers. We can map this to the discrete 6-label scale by associating the labels \emph{pants on fire},  \emph{false}, \emph{barely true}, \emph{half true}, \emph{mostly true}, and \emph{true} with the intervals $[0, \frac{1}{6}), [\frac{1}{6}, \frac{2}{6}), [\frac{2}{6}, \frac{3}{6}), [\frac{3}{6}, \frac{4}{6}), [\frac{4}{6}, \frac{5}{6}), [\frac{5}{6}, 1]$, respectively. We call this method \textbf{question aggregation}. We use the 50 claims and the corresponding questions from the \textbf{Validation-sub} set for evaluation. We also establish the upper bound (\textbf{question aggregation*}) for this heuristic by having one of the authors remove unrelated questions. On average, 0.3 questions are removed  per claim.



Table~\ref{tab:annotation-analysis-label-aggregation} compares our heuristics with simple baselines (random assignment and most frequent class assignment). Our heuristic easily outperforms the baselines, with the predicted label on average is only shifted by one label, e.g., \emph{mostly true} vs.~\emph{true}. This demonstrates the potential of building a more complex model to aggregate subquestion-answer sets, which we leave as a future direction.

Our simple aggregation suffers in the following cases: (1) The subquestions are not equal in importance. The first example in Figure~\ref{fig:annot-analysis} contains two yes subquestions and two no subquestions, and our aggregation yields \emph{half-true} label, differing from gold label \emph{barely-true}. (2) Not all questions are relevant. As indicated by \textbf{question aggregation*}, we are able to achieve better performance after removing unrelated questions. (3) In few cases, the answer to a question could inversely correlate with the veracity of a claim. For example, the claim states "Person X implied Y" and the question asks "Did person X not imply Y?" We think all of the cases can be potentially fixed by stronger models. For example, a question salience model can mitigate (1) and (2), and promotes researches about understanding core arguments of a complex claim. We leave this as future work.


\section{Evidence Retrieval with Decomposition} \label{sec:evidence-retrieval}

Lastly, we explore using claim decomposition for retrieving evidence paragraphs to verify claims. Retrieval from the web to check claims is an extremely hard problem \cite{singh-et-al-2021}. We instead explore a simplified proof-of-concept setting: retrieving relevant paragraphs from the full justification document. These articles are lengthy, containing an average of 12 paragraphs, and with distractors due to entity and concept overlap with the claims. 

We aim to show two advantages of using the decomposed questions: (1) The implied questions contain information helpful to retrieve evidence beyond the lexical information of the claim. (2) We can convert the subquestions to statements and treat them as hypotheses to apply the off-the-shelf NLI models to retrieve evidence that entails such hypotheses~\cite{chen-etal-2021-nli-models}. 

\paragraph{Evidence Paragraph Collection}

We first collect human annotation to identify relevant evidence paragraphs. Given the full PolitiFact verification article consisting of $m$ paragraphs $\mathbf{p} = (p_1,\ldots,p_m)$ and a subquestion, annotators find paragraphs relevant to the subquestion. As this requires careful document-level reading, we hire three undergraduate linguistics students as annotators. We use the 50 claims from the Validation-sub set and present the annotators with the subquestions and the articles. For each subquestion, for each paragraph in the article, we ask the annotators to choose whether it served as context to the subquestion or whether it supports/refutes the subquestion. The statistics and inter-annotator agreement is shown in Table~\ref{tab:evidence-retrieval-annotation-stats}. Out of 12.4 paragraphs on average, 3-4 paragraphs were directly relevant to the claim and the rest of paragraphs mostly provide context.

\begin{table}[t]
\small
\centering
\begin{tabular}{ l   c c }
\toprule
& per subquestion & per example (claim) \\
\midrule
 avg \# of paras & 12.4 & 12.4 \\
 \% of context & 87.6 & 68.8 \\
 \% of support & \phantom{0}5.4 & 12.0 \\ 
 \% of refute & \phantom{0}8.0 & 19.2 \\ 
Fleiss Kappa & 0.42 & 0.42 \\
\bottomrule
\end{tabular}
\caption{Evidence paragraph retrieval data statistics on Validation-sub dataset (50 claims).}
\label{tab:evidence-retrieval-annotation-stats}
\end{table}

\begin{figure}[t]
\centering \vspace{-0.4em}
\includegraphics[width=0.4\textwidth]{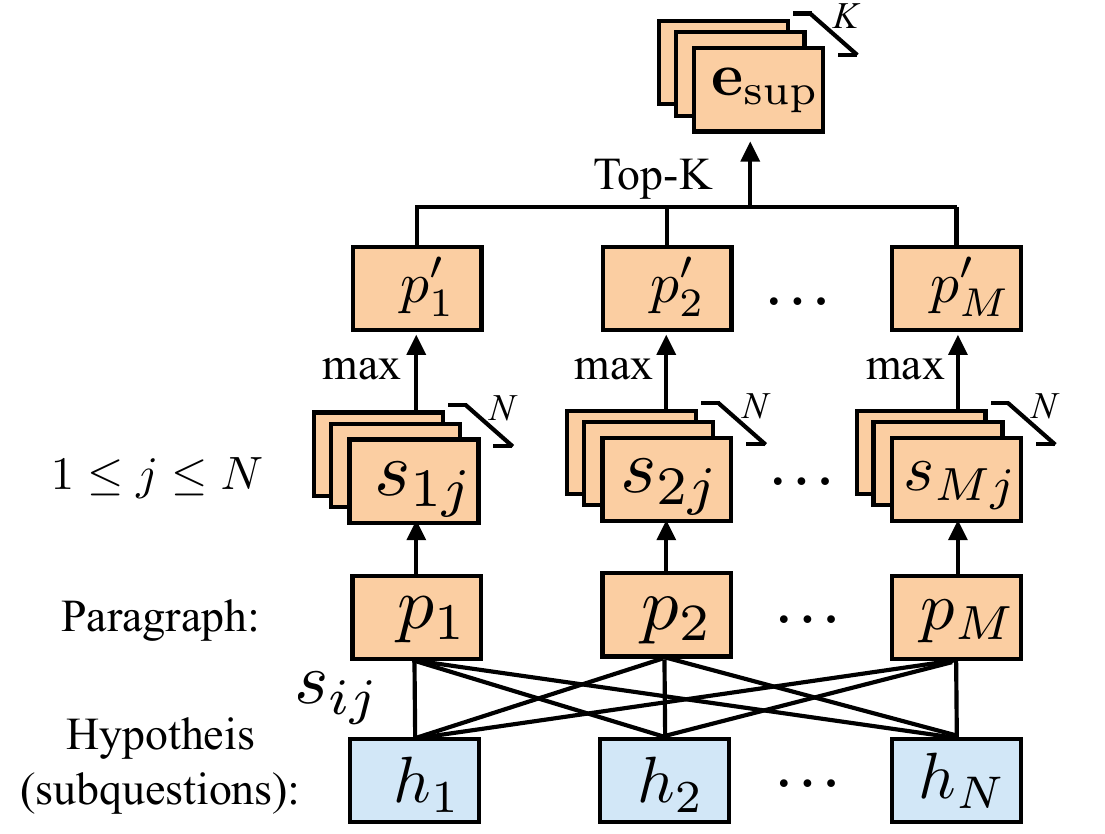}
\caption{Illustration of evidence paragraph retrieval process. The notations corresponds to our descriptions in Section~\ref{sec:evidence-retrieval}. $K$ is a hyperparameter controlling the number of passages to retrieve. }
    \label{fig:evidence-retriever}
\end{figure}

\paragraph{Experimental Setup} 

We experiment with three off-the-shelf RoBERTa-based~\cite{liu2020roberta} NLI models trained on three different datasets: MNLI~\cite{williams-etal-2018-broad}, NQ-NLI~\cite{chen-etal-2021-nli-models}, and DocNLI~\cite{yin-etal-2021-docnli}. We compare the performance of NLI models with random, BM25, and human baselines.



\begin{table}[t]
\small
\centering
\begin{tabular}{ l c  c | c}
\toprule
Model & \multicolumn{2}{c|}{Decomposed claim} & Original \\
 & predicted & gold  &  claim  \\
\midrule
  MNLI  & 41.0  &  48.8 & 35.2 \\ 
 NQ-NLI  & 38.8 &  34.5 & 40.9 \\
 DocNLI  & \textbf{44.7} & \textbf{59.6} & 36.9 \\
 BM25 & 36.2 & 47.5 & 39.2 \\
\bottomrule
\end{tabular}
\caption{Evidence retrieval performance (F1 score) with the decomposed claims (from predicted and annotated (gold) subquestions) and the original claim on the Validation-sub set. A random baseline achieves 24.9 F1 and human annotators achieve 69.0 F1.}
\vspace{-0.3cm}
\label{tab:exp-NLI-retrieval}
\end{table}

We first convert the corresponding subquestions $\mathbf{q} = {q_1, ..., q_N}$ of claim $c$ to a set of statements $\mathbf{h} = {h_1, ..., h_n}$ using GPT-3~\cite{brown2020language}.\footnote{We release the automatically converted statements and the negations for all of the subquestions in the published dataset.} We find that with only 10 examples as demonstration, GPT-3 can perform the conversion quite well (with an error rate less than 5\%). For more information about the prompt see Appendix~\ref{sec:GPT3-prompt}. 

To retrieve the evidence that \textbf{supports} the statements, we treat the statements as hypotheses and the paragraphs in the article as premises. We feed them into an NLI model to compute the score associated with the ``entailment'' class for every premise and hypothesis pair. Here, the score for paragraph $p_i$ and hypothesis $h_j$ is defined as the output probability $s_{ij} = P(\text{Entailment} \mid  p_i, h_j)$. We then select as evidence the top $k$ paragraphs by score across all subquestions: for paragraph $p_i$, we define $p_i' = \max (\{s_{ij} \mid 1 \leq j \leq N\})$, which denotes for each hypothesis from 1 to $N$ that the $j$th hypothesis $h_j$ achieves the highest score with $p_i$. Then $\mathbf{e_{sup}} = \{ p_i \mid i \in \text{Top-K}(\{p'_1, ..., p'_M\})\}$. We set $k$ to be the number of the paragraphs that are annotated with either support or refute.  Figure~\ref{fig:evidence-retriever} describes this approach.

To retrieve the evidence that \textbf{refutes} the statements, we follow the same process, but with the negated hypotheses set $\mathbf{h}$ generated by GPT3. (Note that our NLI models trained on NQ-NLI and DocNLI only have two classes, entailed and not entailed, and not entailed is not a sufficient basis for retrieval.) The final evidence set is obtained by merging the evidence from the \emph{support} and \emph{refute} set. This is achieved by removing duplicates then taking Top-K paragraphs according to the scores.

\textbf{BM25 baseline model} uses retrieval score instead of NLI score. \textbf{The random baseline} randomly assign support, refute, neutral labels to paragraphs based on the paragraph label distribution in Table~\ref{tab:evidence-retrieval-annotation-stats}. \textbf{Human performance} is computed by selecting one of the three annotators and comparing their annotations with the other two (we randomly pick one annotator if they do not agree), taking the average over all three annotators. This is not directly comparable to the annotations for the other techniques as the gold labels are slightly different.
\paragraph{Results} 

The results are shown in Table~\ref{tab:exp-NLI-retrieval}. We see that \textbf{the decomposed questions are effective to retrieve the evidence.} By aggregating evidence from the subquestions, both BM25 and the NLI models can do better than using the claim alone, except for the case of using DocNLI, and BM25 with the predicted decomposition. The best model with gold annotations (59.6) is close to human performance (69.0) in this limited setting, indicating that the detailed and implied information in decomposed questions can help gathering evidence beyond the surface level of the claim. 

\textbf{DocNLI outperforms BM25 on both the annotated decomposition and the predicted decomposition.} This demonstrates the potential of using the NLI models to aid the evidence retrieval in the wild, although they must be combined with decomposition to yield good results. 




\section{Related Work}
\paragraph{Fact-checking} 
~\citet{vlachos-riedel-2014-fact} proposed to decompose the fact-checking process into three components: identifying check-worthy claims, retrieving evidence, and producing verdicts. Various datasets have been proposed, including human-generated claims based on Wikepedia~\cite{thorne-etal-2018-fever,chen2019tabfact, jiang-etal-2020-hover, schuster-etal-2021-get, aly2021feverous}, real-world political claims~\cite{wang-2017-liar, alhindi-etal-2018-evidence, augenstein-etal-2019-multifc, Ostrowski2021MultiHopFC, gupta-srikumar-2021-x}, and science claims~\cite{wadden-etal-2020-fact, saakyan-etal-2021-covid}. Our dataset focuses on real-world political claims, particularly more complex claims than past work which necessitate the use of decompositions. 

Our implied subquestions go beyond what is mentioned in the claim, asking the intention and political agenda of the speaker. \citet{Gabriel2022MisinfoRF} study such implications by gathering expected readers' reactions and writers' intentions towards news headlines, including fake news headlines.

To produce verdicts of the claims, other work generates explanations for models' predictions. ~\citet{popat2017truth, popat-etal-2018-declare, shu2019defend,yang2019xfake, lu-li-2020-gcan} presented attention-based explanations; ~\citet{gad2019exfakt,ahmadi2019explainable} used logic-based systems, and ~\citet{atanasova-etal-2020-generating-fact, kotonya-toni-2020-explainable-automated} modeled the explanation generation as a summarization task. Combining answers to the decomposed questions in our work can form an explicit explanation of the answer. 

\paragraph{Question Generation}
Our work also relates to question generation (QG)~\cite{du-etal-2017-learning}, which has been applied to augment data for QA models~\cite{duan-etal-2017-question, sachan-xing-2018-self, alberti-etal-2019-synthetic}, evaluate factual consistency of summaries~\cite{wang-etal-2020-asking, durmus-etal-2020-feqa,Kamoi-Et-Al2022}, identify semantic relations ~\cite{he-etal-2015-question,klein-etal-2020-qanom,pyatkin-etal-2020-qadiscourse}, and identify useful missing information in a given context (clarification)~\cite{rao-daume-iii-2018-learning, shwartz-etal-2020-unsupervised, majumder-etal-2021-ask}. Our work is most similar to QABriefs~\cite{fan-etal-2020-generating}, but differs from theirs in two ways: (1) We generate yes-no questions directly related to checking the veracity of the claim. (2) Our questions are more comprehensive and precise. 

\section{Conclusion}
We present a dataset containing more than 1,000 real-world complex political claims with their decompositions in question form. With the decompositions, we are able to check the explicit and implicit arguments made in the claims. We also show the decompositions can play an important role in both evidence retrieval and veracity composition of an explainable fact-checking system. 

\section{Limitations} \label{sec:limitations}

    \paragraph{Interaction of retrieval and decomposition} The evidence retrieval performance depends on the quality of the decomposed questions (compare our results on generated questions to those on annotated questions in Section~\ref{sec:evidence-retrieval}). Yet, generating high-quality questions requires relevant evidence context. These two modules cannot be strictly pipelined and we envision that in future work, they will need to interact in an iterative fashion. For example, we could address this with a human-in-the-loop approach. First, retrieve some context passages with the claim to verify as a query, possibly focused on the background of the claim and the person who made the claim. This retrieval can be done by a system or a fact-checker. Then, we use context passages to retrain the QG model with the annotations we have and the fact-checker can make a judgment about those questions, adding new questions if the generated questions do not cover the whole claim. We envision that such a process can make fact-checking easier while providing data to train the retrieval and QG models.

\paragraph{Difficulty of automatic question comparison}
As discussed in section~\ref{sec:auto_decomp}, automatic metrics to evaluate our set of generated questions do not align well with human judgments. Current automatic metrics are not sensitive enough to minor changes that could lead to different semantics for a question. For example, changing \emph{``Are \textbf{all} students in Georgia required to attend chronically failing schools?''} to \emph{``Are students in Georgia required to attend chronically failing schools?''} yields two questions that draw an important contrast. However, we will get an extremely high BERTScore (0.99) and ROUGE-L score (0.95) between the two questions. Evaluating question similarity without considering how the questions will be used is challenging, since we do not know what minor distinctions in questions may be important. We suggest measuring the quality of the generated questions on some downstream tasks, e.g., evidence retrieval.   

\paragraph{General difficulty of the task} We have not yet built a full pipeline for fact-checking in the true real-world scenario. Instead, we envision our proposed question decomposition as an important step of such a pipeline, where we can use the candidate decompositions to retrieve deeper information and verify or refute each subquestion, then compose the results of the subquestions into an inherently explainable decision. In this paper, we have shown that the decomposed questions can help the retriever in a clean setting. But retrieving evidence in the wild is extremely hard since some statistics are not accessible through IR and not all available
information is trustworthy \cite{singh-et-al-2021}, which are issues beyond the scope of this paper. Through the \textbf{Question Aggregation} probing, we also show the potential of composing the veracity of claims through the decisions from the decomposed questions. The proposed dataset opens a door to study the core argument of a complex claim.

\paragraph{Domain limitations and lack of representation} The dataset we collected only consists of English political claims from the PolitiFact website. These claims are US-centric and largely focused on politics; hence, our results should be interpreted with respect to these domain limitations.

\paragraph{Broader impact: opportunities and risks of deployment} Automated fact checking can help prevent the propagation of misinformation and has great potential to bring value to society. However, we should proceed with caution as the output of a fact-checking system---the veracity of a claim---could alter users' views toward someone or something. Given this responsibility, we view it as crucial to develop explainable fact-checking systems which inform the users which parts of the claim are supported, which parts are not, and what evidence supports these judgments. In this way, even if the system makes a mistake, users will be able to check the evidence and draw the conclusion themselves.

Although we do not present a full fact-checking system here, we believe our dataset and study can help pave the way towards building more explainable systems. By introducing this claim decomposition task and the dataset, we will enable the community to further study the different aspects of real-world complex claims, especially the implied arguments behind the surface information.

\section*{Acknowledgments}
This dataset was funded by Good Systems,\footnote{\url{https://goodsystems.utexas.edu/}} a UT Austin Grand Challenge to develop responsible AI technologies, as well as NSF Grant IIS-1814522, NSF CAREER Award IIS-2145280, and gifts from Salesforce and Adobe. We would like to thank our annotators from Upwork and UT Austin: Andrew Baldino, Scarlett Boyer, Catherine Coursen, Samantha Dewitt, Abby Mortimer, E Lee Riles, Keegan Shults, Justin Ward, Meona Khetrapal, Misty Peng, Payton Wages, and Maanasa V Darisi.

\bibliography{anthology,custom}
\bibliographystyle{acl_natbib}

\appendix

\section{Question Annotation Workflow} \label{sec:question-annotation-workflow}
\subsection{Workflow}
Tracing the thought process of professional fact-checkers requires careful reading. Thus, instead of using crowdsourcing platforms with limited quality control, we recruit 8 workers with experience in literature or politics from the freelancing platform Upwork.\footnote{\url{https://www.upwork.com/}} We pay the annotators \$1.75 per claim, which translates to around \$30/hour.\footnote{We asked the workers to report their speeds at the end of the task and found their actual hourly rates ranged between \$18 and \$50 per hour.} Each annotator labeled an initial batch of articles and we provided feedback on their annotation. We communicated with annotators during the process. 

We posted a job advertisement including the description and the payment plan of our task on the Upwork platform. In total 14 workers applied for the position. We first conducted an initial qualification round in which we released an initial batch of 15 documents for the annotators to complete, for which we paid \$35. This initial batch was used to judge how suitable the annotators are for this task. We reviewed the annotations and give detailed feedback to each annotator for every claim along with our suggested annotation for reference. We selected annotators whose annotation met our qualifications to continue to the next round. In the initial round, we selected 8 out of the 14 annotators who applied. 

After the initial round, we released new example batches to the annotators on a weekly basis. Each batch contained 100 examples for which we paid \$175. The hired annotators were required to complete at least one batch per week and they could do up to 2 batches per week.\footnote{We intentionally limited this to avoid having a single annotator annotate a large portion of the examples.}

\subsection{Annotation Interface}
\begin{figure*}[t]
\centering
\includegraphics[width=\textwidth]{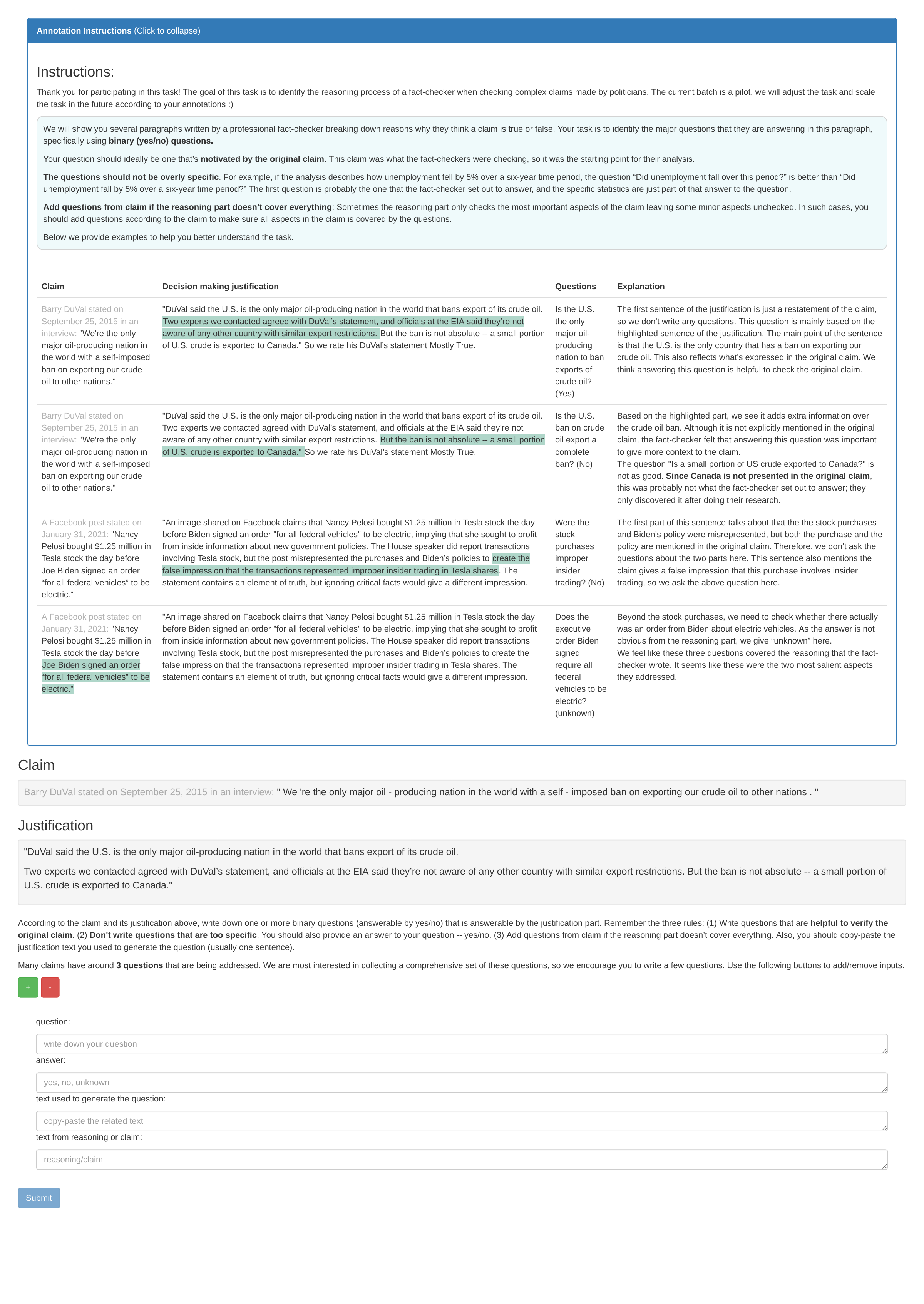}
\caption{Interface of our question decomposition task, including the annotation instructions.}
    \label{fig:question-annotation-interface}
\end{figure*}

The interface of the main question decomposition task is shown in Figure~\ref{fig:question-annotation-interface}. 

\section{Evidence Annotation Interface} \label{sec:evidence-annotation-interface}
\begin{figure*}[t]
\centering
\includegraphics[width=\textwidth]{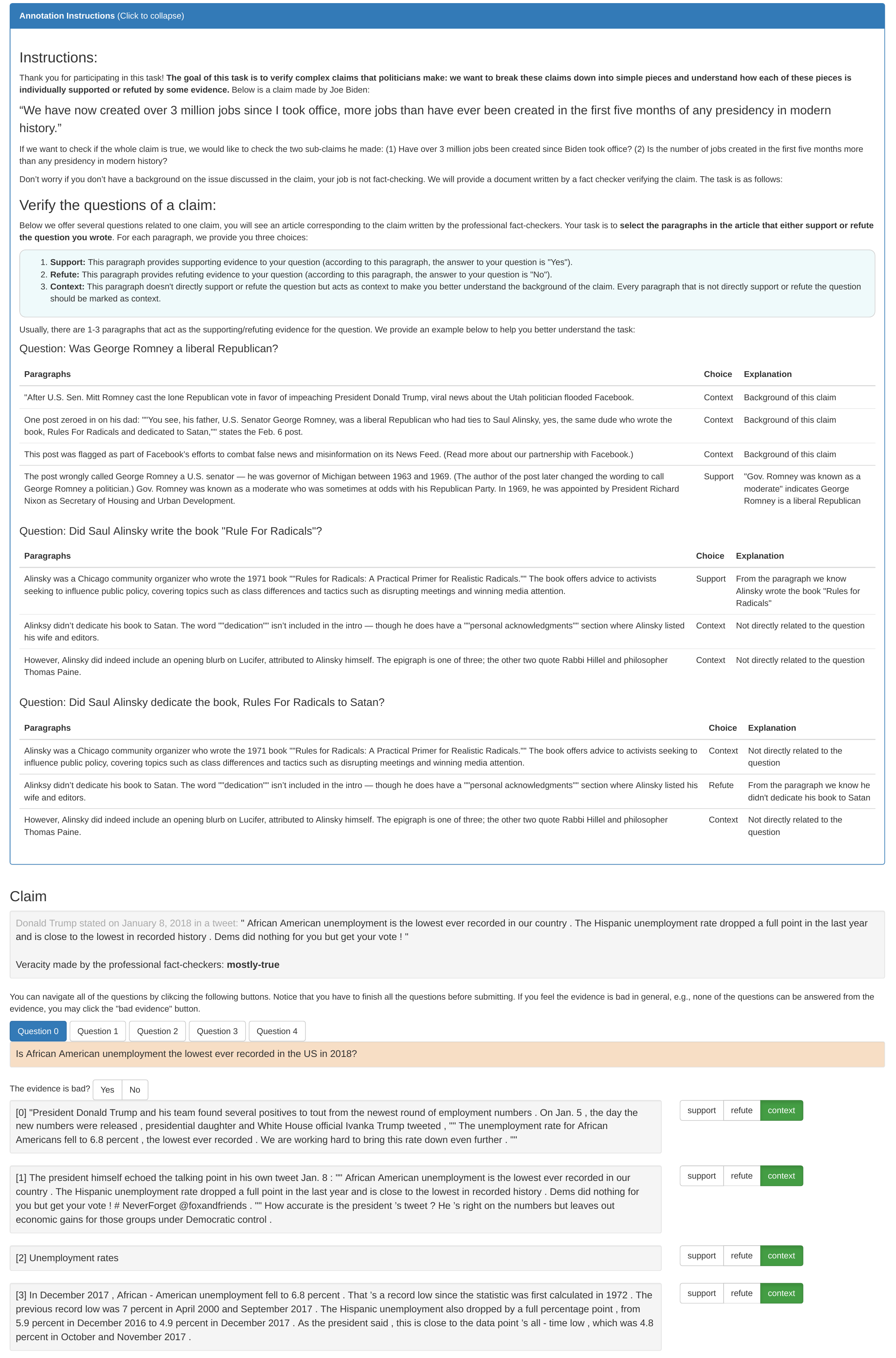}
\caption{Interface of our evidence annotation task used in section~\ref{sec:evidence-retrieval}, including the annotation instructions.}
    \label{fig:evidence-annotation-interface}
\end{figure*}

The interface to annotate the supporting/refuting evidence described in section~\ref{sec:evidence-retrieval} is shown in Figure~\ref{fig:evidence-annotation-interface}. 

\section{User Study Interface} \label{sec:user-study-interface}
\begin{figure*}[t]
\centering
\includegraphics[width=\textwidth]{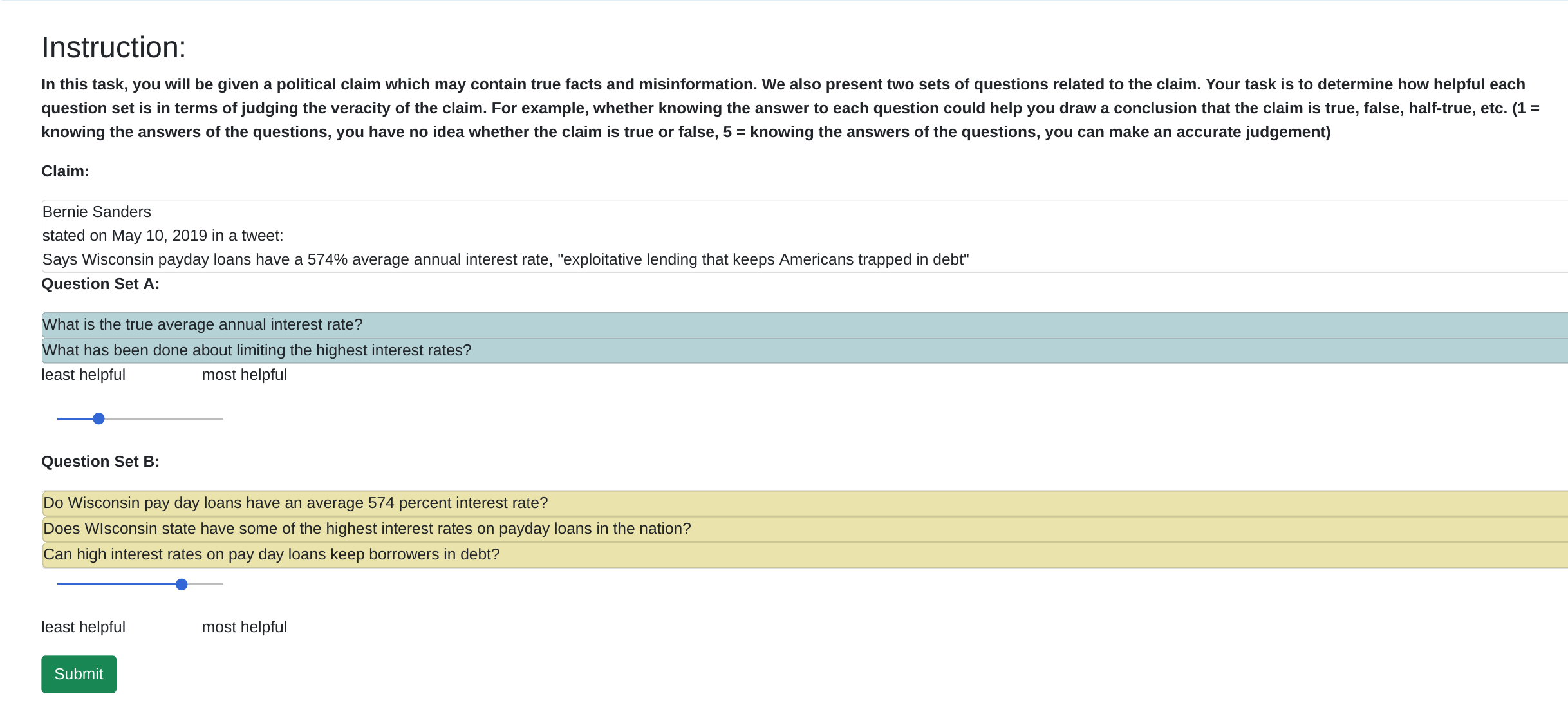}
\caption{Interface of our user study conducted in section~\ref{sec:question-comparison}, including the annotation instructions.}
    \label{fig:question-comparison-interface}
\end{figure*}

The annotation interface of our user study conducted in Section~\ref{sec:question-comparison} is shown in Figure~\ref{fig:question-comparison-interface}.

\section{Inter-annotator Agreement} \label{sec:inter-annotator-agreement-example}
\begin{figure*}[t]
\centering
\includegraphics[width=\textwidth]{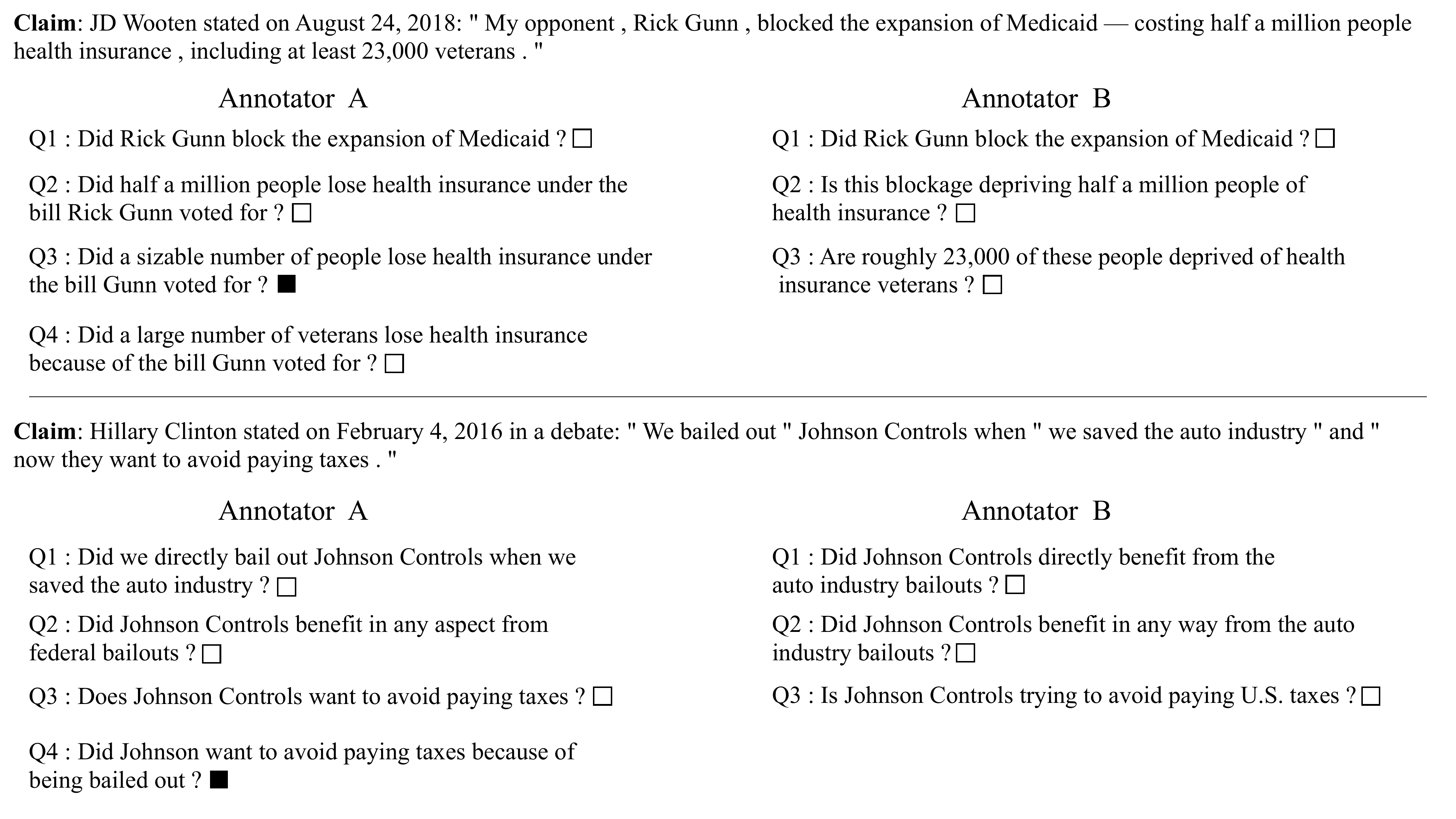}
\caption{Two examples of our inter-annotator agreement assessment: giving two set of questions, we mark questions of which the semantics cannot be matched to the other annotator’s set. Here, the black box denotes the marked question.}
    \label{fig:unique-question-annotation}
\end{figure*}

Two examples of our inter-annotator agreement assessment are shown in Figure~\ref{fig:unique-question-annotation}. In the first example, we treat Q3 of annotator A as not covered by annotator B. It is a weaker version of Q2 but not mentioned by annotator B. Q4 of annotator A has similar semantics as Q3 of annotator B so we do not mark it.

\section{Automatic Claim Decomposition Evaluation}
\label{sec:claim_decomp_eval}


For the evaluation in Section~\ref{sec:auto_decomp}, we also explored an automated method for assessing whether generated questions match ground truth ones. We aim to define a metric $m(\mathbf{q},\mathbf{\hat{q}})$ that compares the two sets of generated questions. However, we lack good off-the-shelf methods for comparing sets of strings like this. Instead, we rely on existing scoring functions that can compare single strings, like ROUGE and BERTScore~\cite{zhang2019bertscore}. 

Following other alignment-based methods like SummaC~\cite{laban2021summac} for summarization factuality, we view these metrics as:
$m(\mathbf{q},\mathbf{\hat{q}}) = \textrm{argmax}_\mathbf{a} \sum s(q_{a_i},\hat{q_i})$, where $\mathbf{a}$ is an alignment variable. This problem can be viewed as finding the maximum-weight matching in a bipartite graph. We use the Hungarian algorithm~\cite{kuhn1955hungarian} to compute this alignment and we take the mean of max matching as the result. The results are shown in Table~\ref{tab:qg-eval-automatic}. 

\paragraph{The automatic metrics are not well aligned with the human judgments.} We see that the Pearson coefficient between human judgments and the automatic metrics ranges from 0.42--0.54 and 0.21--0.45 for \textsc{qg-multiple} and \textsc{qg-nucleus} respectively. The large instability of the Pearson coefficient  indicates that the automatic evaluation may not accurately reflect the quality of the generated questions. Therefore, evaluating the generated questions on downstream tasks could be more accurate, hence why we also study evidence retrieval.

\begin{table*}[t]
\small
\centering
\begin{tabular}{ l | c c c c}
\toprule
Model & Rouge-1 (P) & Rouge-2 (P) & Rouge-l (P) & Bert-score (P) \\
\midrule
\textsc{qg-multiple} &  0.44 (0.54) &	0.23 (0.47) &  0.40 (0.53)	& 0.92 (0.42)  \\
\textsc{qg-nucleus} &  0.39 (0.32) & 0.18 (0.21) & 0.36 (0.32) & 0.91 (0.45) \\
\midrule
\textsc{qg-multiple-justify} & 0.54 (0.36) & 0.38 (0.35) & 0.52 (0.37) & 0.93 (0.35)\\
\textsc{qg-nucleus-justify} & 0.41 (0.25) & 0.20 (0.35) & 0.37 (0.30) & 0.91 (0.41) \\
\bottomrule
\end{tabular}
\caption{Automatic evaluation results on the development set. Here, (P) denotes the Pearson correlation coefficient between the automatic metric and recall-all. \textsc{-justify} denotes training the question generator by concatenating the claim and the justification paragraph as the input.
 }
\label{tab:qg-eval-automatic}
\end{table*}

\section{Training Details for Question Generation} \label{sec:model-details}

For \textsc{qg-multiple}, each instance of the input and the output are constructed according to the template:
$$N \text{[S]} \text{c} \text{[S]} \longrightarrow \text{q}_1 \text{[S]} \text{q}_2 \text{[S]}, ...,  \text{q}_N$$
where [S] denotes the separator token of the T5 model. $N$ denotes the number of questions to generate; we introduce this into the input to serve as a control variable and set it to match the number of annotated questions during training. $c$ denotes the claim and $\text{q}_i$ denotes the $i$th annotated question and we do not assume a specific order for the questions.

The model is trained using the seq2seq framework of Hugging Face~\cite{wolf-etal-2020-transformers}. The max sequence length for input and output is set to 128 and 256 respectively. The batch size is set to 8 and we use DeepSpeed for memory optimization~\cite{rasley2020deepspeed}. We train the model on our training set for 20 epochs with AdamW~\cite{loshchilov2018decoupled} optimizer and an initial learning rate set to 3e-5. 

At inference, we use beam search with beam size set to 5. We prepend the number of questions to generate ($N$) at the start of the claim in the input.

For \textsc{qg-nucleus}, we construct multiple input-output instances $(c \rightarrow q_i)$ for each claim, where $q_i$ denotes the $i$th decomposed question of claim $c$. The max sequence length for input and output are both set to 128. The batch size is set to 16 and we use DeepSpeed for memory optimization. We train the model on our training set for 10 epochs with AdamW optimizer and an initial learning rate set to 3e-5.  

We expect this model to place a flatter distribution over the output space, assigning many possible questions high weight due to the training data including multiple outputs for the same input. At inference, we use nucleus sampling~\cite{holtzman2019curious} in which $p$ is set to 0.95 together with top-$k$ sampling~\cite{fan-etal-2018-hierarchical} in which $k$ is set to 50 to generate questions. We filter out the duplicates (exact string match) in the sampled questions set.

\section{GPT-3 for Question Conversion} \label{sec:GPT3-prompt}
Given a question, we let GPT-3 generate its declarative form as well as the negated form of the statement. We achieve this by separating them using ``|'' in the prompt. One advantage of using GPT-3 is that it can easily generate natural sentences. For example, for question \emph{"Are any votes illegally counted in the election?"}, GPT-3 generates the statement and its negation as \emph{"Some votes were illegally counted in the election."} and \emph{"No votes were illegally counted in the election."}. A demonstration of the prompt we used for the question conversion is shown as follows:\\ \\
\begin{minipage}{.45\textwidth}
\fontfamily{Cabin}
\selectfont
\small

Question: Are unemployment rates for African Americans and Hispanics low today?\\
Statement: Unemployment rates for African Americans and Hispanics are low today. | Unemployment rates for African Americans and Hispanics are not low today. \\

Question: Were 1700 mail-in ballots investigated for fraud in Texas?\\
Statement: 1700 mail-in ballots were investigated for fraud in Texas | 1700 mail-in ballots were not investigated for fraud in Texas \\

Question: Is Wisconsin guaranteeing Foxconn nearly \$3 billion?\\
Statement: Wisconsin guarantees Foxconn nearly \$3 billion. | Wisconsin does not guarantee Foxconn nearly \$3 billion.\\

Question: Will changes in this law raise taxes for anyone? \\
Statement: The changes in this law will raise taxes for someone. | The changes in this law will raise taxes for no one. \\

Question: Has Donnelly directly sponsored any of these legislative proposals since becoming a senator? \\
Statement: Donnelly directly sponsored some of these legislative proposals since becoming a senator. | Donnelly directly sponsored none of these legislative proposals since becoming a senator. \\

... \\ 

Question: INPUT-QUESTION \\
Statement: MODEL-OUTPUT \\
\end{minipage}

\section{Qualitative Analysis of Generated Questions} \label{sec:qualitative-analysis-qg}

\begin{figure*}[t]
\centering
\includegraphics[width=\textwidth]{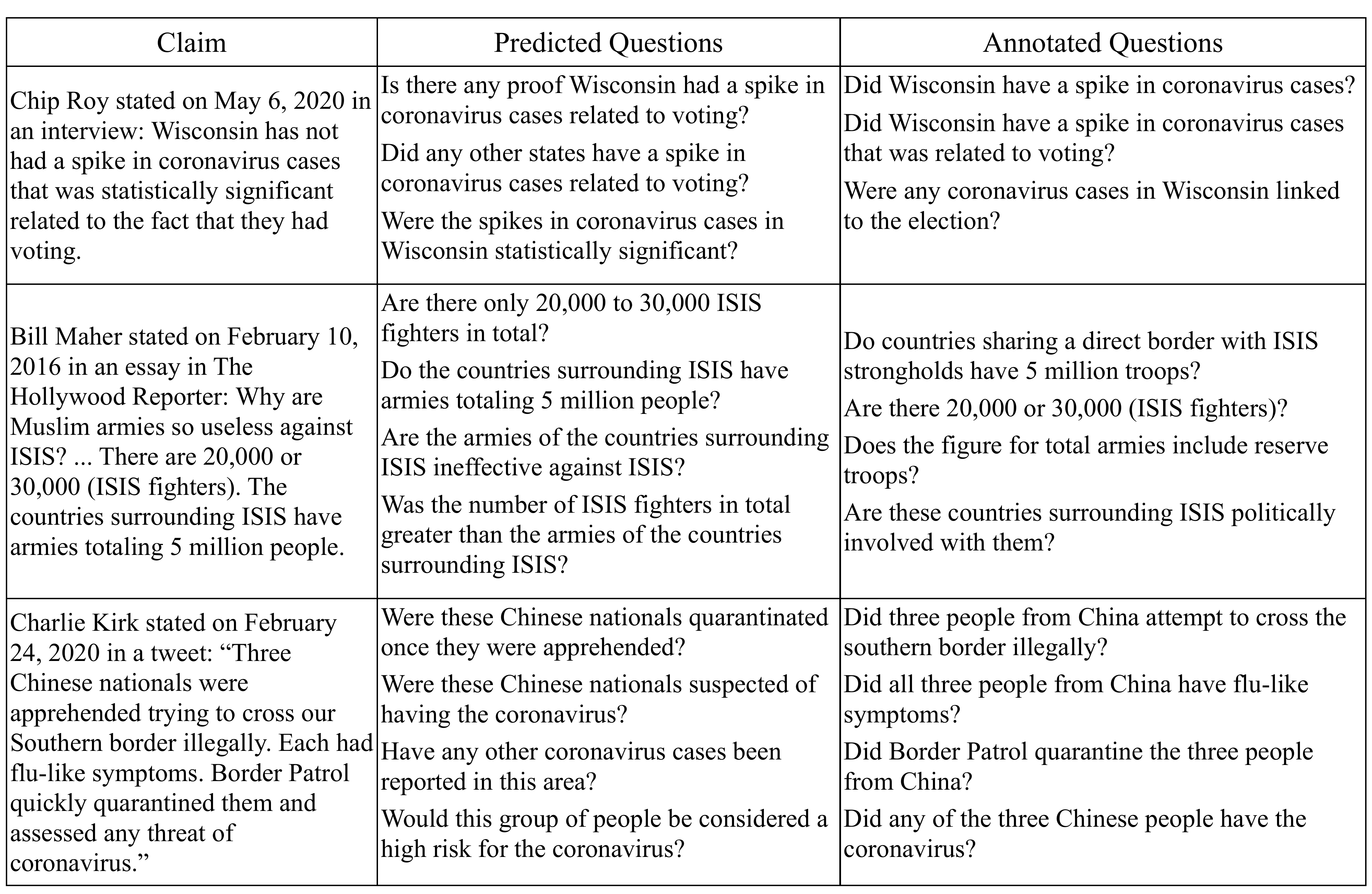}
\caption{Questions generated by the \textsc{qg-multiple} model, compared with the annotations.}
    \label{fig:qualitative-analysis-qg}
\end{figure*}

Table~\ref{fig:qualitative-analysis-qg} includes more examples where the generated questions do not match the annotations but also worth checking. For example, for the second claim, our model generates the question \emph{``Did any other states have a spike in coronavirus cases related to voting?''} Although the gold fact-check did not address this question, this kind of context is the kind of thing a fact-checker may want to be attentive to, even if the answer ends up being no, and we judge this to be a reasonable question to ask given only the claim.

\section{More examples of QABriefs} \label{sec:compare-to-fan-et-al-more-examples}
\begin{figure*}[t]
\centering
\includegraphics[width=\textwidth]{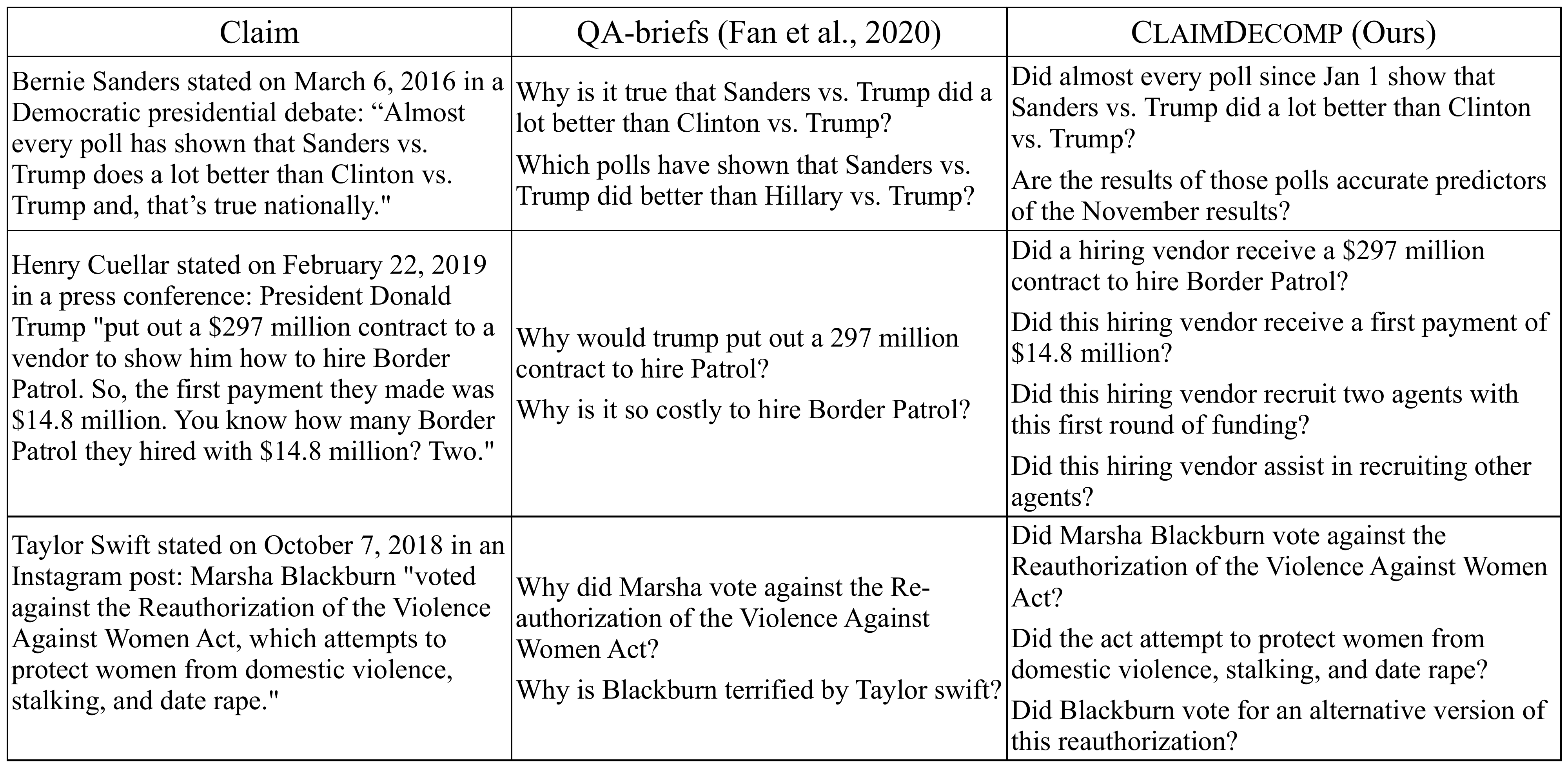}
\caption{More examples of the comparison between our decomposed questions with QABriefs~\cite{fan-etal-2020-generating}.}
    \label{fig:annot-analysis-vs-fan-et-al-more-examples}
\end{figure*}

We include more examples reflecting the annotation difference between our method and QABriefs in Figure~\ref{fig:annot-analysis-vs-fan-et-al-more-examples}.

\section{Datasheet for \dataname}\label{sec:datasheet}
\subsection{Motivation for Datasheet Creation}
\paragraph{Why was the dataset created?}
Despite the progress made in automating the fact-checking process, the performance achieved by current models is relatively poor. Systems in this area fundamentally need to be designed with an eye towards human verification, motivating our effort to build more explainable models so that the explanations can be used to interpret a model's behavior. Therefore, we create this dataset to facilitate future research to achieve this goal. We envision that by verifying each question, we can compose the final veracity of the claim in inherently explainable way.

\paragraph{Has the dataset been used already?}
The dataset has not been used beyond the present paper, where it was used to train a question generation model and in several evaluation conditions.

\paragraph{Who funded the dataset?}
This dataset was funded by Good Systems,\footnote{\url{https://goodsystems.utexas.edu/}} a UT Austin Grand Challenge to develop responsible AI technologies.

\subsection{Dataset Composition}
\paragraph{What are the instances?}
Each instance is a real-world political claim. All claims are written in English and most of them are US-centric.

\paragraph{How many instances are there?}
Our dataset consists of two-way annotation of 1,200 claims, and 6,555 decomposed questions. A detailed breakdown of the number of instances can be seen in Table~\ref{tab:dataset-statistics} of the main paper.

\paragraph{What data does each instance consist of?}
Each instance contains a real-world political claim and a set of yes-no questions with associated answers.

\paragraph{Does the data rely on external resources?}
Yes, assembling it requires access to PolitiFact.

\paragraph{Are there recommended data splits or evaluation measures?}
We include the recommended train, development, and test sets for our datasets. The distribution can be found in Table~\ref{tab:dataset-statistics}.

\subsection{Data Collection Process}
\paragraph{How was the data collected?}
We recruit 8 annotators with background in literature or politics from the freelancing platform Upwork. Given a claim paired with the justification written by the professional fact-checker on PolitiFact, we ask our annotators to reverse engineer the fact-checking process: generate yes-no questions which are answered in the justification part. For each question, the annotators also give the answer and select the relevant text in the justification that is used for the generation. The annotators are instructed to cover as many of the assertions made in the claim as possible without being overly specific in their questions.

\paragraph{Who was involved in the collection process and what were their roles?}
The 8 annotators we recruited perform the all the annotation steps outlined above.

\paragraph{Over what time frame was the data collected?}
The dataset was collected over a period from January to April 2022.

\paragraph{Does the dataset contain all possible instances?}
Our dataset does not cover all possible political claims. It mainly include complex political claims made by notable political figures of the U.S. through 2012 to 2021.

\paragraph{If the dataset is a sample, then what is the population?}
It represents a subset of all possible complex political claims which require verifying multiple aspects of the claim to reach a final veracity. Our dataset also only includes claims written in English.

\subsection{Data Preprocessing}

\paragraph{What preprocessing / cleaning was done?}
We remove any additional whitespace in the annotated questions, but otherwise we do not postprocess the annotations in any way. 

\paragraph{Was the raw data saved in addition to the cleaned data?}
Yes

\paragraph{Does this dataset collection/preprocessing procedure achieve the initial motivation?}
Our collection process indeed achieves our initial goals of creating a high-quality dataset of complex political claims with the decompositions in question form. Using this data, we are able to check the explicit and implicit arguments made by the politicians.

\subsection{Dataset Distribution}
\paragraph{How is the dataset distributed?}
We make our dataset available at \url{https://jifan-chen.github.io/ClaimDecomp}.

\paragraph{When was it released?}
Our data and code is currently available.

\paragraph{What license (if any) is it distributed under?}
\dataname{} is distributed under the CC BY-
SA 4.0 license.\footnote{\url{https://creativecommons.org/licenses/by-sa/4.0/legalcode}}

\paragraph{Who is supporting and maintaining the dataset?}
This dataset will be maintained by the authors of this paper. Updates will be posted on the dataset website.

\subsection{Legal and Ethical Considerations}

\paragraph{Were workers told what the dataset would be used for and did they consent?}
Crowd workers informed of the goals we sought to achieve through data collection. They also consented to have their responses used in this way through the Amazon Mechanical Turk Participation Agreement (note that even though we recruited through Upwork, workers performed annotation in the Mechanical Turk sandbox).

\paragraph{If it relates to people, could this dataset expose people to harm or legal action?}
Our dataset does not contain any personal information of crowd workers. However, our dataset can include incorrect information in the form of false claims. These claims were made in a public setting by notable political figures; in our assessment, such claims are already notable and we are not playing a significant role in spreading false claims as part of our dataset. Moreover, these claims are publicly available on PolitiFact along with expert assessment of their correctness. We believe that there is a low risk of someone being misled by information they see presented in our dataset.

\paragraph{If it relates to people, does it unfairly advantage or disadvantage a particular social group?}
We acknowledge that, because our dataset only covers English and annotators are required to be located in the US, our dataset lacks representation of claims that are relevant in other languages and to people around the world. The claims themselves could reflect misinformation rooted in racism, sexism, and other forms of intergroup bias.

\end{document}